\def\year{2020}\relax
\documentclass[letterpaper]{article} 
\usepackage{aaai20}  
\usepackage{times}  
\usepackage{helvet} 
\usepackage{courier}  
\usepackage[hyphens]{url}  
\usepackage{graphicx} 
\usepackage{graphicx}  
\newcommand{\citet}[1]{\citeauthor{#1}~\shortcite{#1}}
\newcommand{\citep}{\cite}

\usepackage{enumitem}

\usepackage{amsmath}
\usepackage{verbatim}

\title{Automatically Neutralizing Subjective Bias in Text }
\author{Anonymous; Paper ID 211}
\author{Reid Pryzant,\textsuperscript{\rm 1} Richard Diehl Martinez,\textsuperscript{\rm 1} Nathan Dass,\textsuperscript{\rm 1}\\ \Large \textbf{Sadao Kurohashi,\textsuperscript{\rm 2} Dan Jurafsky,\textsuperscript{\rm 1} Diyi Yang\textsuperscript{\rm 3}} \\
\textsuperscript{\rm 1}Stanford University \\  \{rpryzant,rdm,ndass,jurafsky\}@stanford.edu \\
\textsuperscript{\rm 2}Kyoto University\\ kuro@i.kyoto-u.ac.jp \\
\textsuperscript{\rm 3}Georgia Institute of Technology\\
diyi.yang@cc.gatech.edu \\
}
 
\begin{document}

\maketitle

\begin{abstract}
Texts like news, encyclopedias, and some social media strive for objectivity. Yet bias in the form of inappropriate subjectivity --- introducing attitudes via framing, presupposing truth, and casting doubt --- remains ubiquitous. This kind of bias erodes our collective trust and fuels social conflict. To address this issue, we introduce a novel testbed for natural language generation: automatically bringing inappropriately subjective text into a neutral point of view (``neutralizing'' biased text). We also offer the first parallel corpus of biased language. The corpus contains 180,000 sentence pairs and originates from Wikipedia edits that removed various framings, presuppositions, and attitudes from biased sentences. Last, we propose two strong encoder-decoder baselines for the task. A straightforward yet opaque \textsc{concurrent} system uses a BERT encoder to identify subjective words as part of the generation process. An interpretable and controllable \textsc{modular} algorithm separates these steps, using (1) a BERT-based classifier to identify problematic words and (2) a novel \emph{join  embedding} through which the classifier can edit the hidden states of the encoder. Large-scale human evaluation across four domains (encyclopedias, news headlines, books, and political speeches) suggests that these algorithms are a first step towards the automatic identification and reduction of bias. 
\end{abstract}

\section{Introduction} 

Writers and editors of texts like encyclopedias, news, and textbooks strive to avoid biased language.  Yet bias remains ubiquitous. 62\% of Americans believe their news is biased \cite{gallup} and bias is the single largest source of distrust in the media \cite{knightSurvey}.

\begin{figure}[]
\includegraphics[width=1\linewidth]{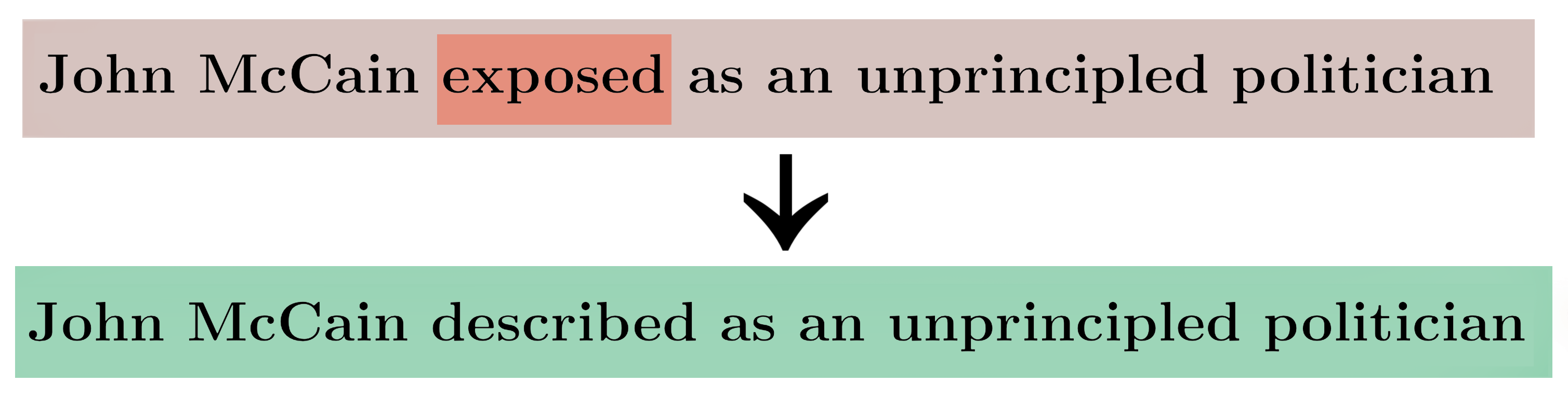}
\caption{Example output from our \textsc{modular} algorithm. ``Exposed'' is a factive verb that presupposes the truth of its complement  (that McCain is unprincipled). Replacing ``exposed'' with ``described'' neutralizes the headline because it conveys a similar main clause proposition (someone is asserting McCain is unprincipled), but no longer introduces the authors subjective bias via  presupposition.}
\label{figure:cnn}
\end{figure}

This work presents data and algorithms for automatically reducing bias in text. We focus on a particular kind  of bias: \emph{inappropriate subjectivity} (``subjective bias''). Subjective bias occurs when language that should be neutral and fair is skewed by feeling, opinion, or taste (whether consciously or unconsciously). In practice, we identify subjective bias via the method of  \citet{recasens2013linguistic}: using Wikipedia's \textit{neutral point of view (NPOV)} policy.\footnote{\url{https://en.wikipedia.org/wiki/Wikipedia:Neutral_point_of_view}} This policy is a set of principles which includes ``avoiding stating opinions as facts'' and ``preferring nonjudgemental language''.

For example a news headline like ``John McCain exposed as an unprincipled politician"  (Figure \ref{figure:cnn}) is biased because the verb \textit{expose} is a factive verb that presupposes the truth of its complement; a non-biased sentence would use a verb like \textit{describe} so as not to presuppose the subjective opinion of the writer.
``Pilfered'' in ``the gameplay is \emph{pilfered} from DDR'' (Table~\ref{tab:samples}) subjectively frames the shared gameplay as a kind of theft.  ``His'' in ``a lead programmer usually spends \emph{his} career'' again introduces a biased and subjective viewpoint (that all programmers are men) through presupposition. 

We aim to debias text by suggesting edits that would make it more neutral. This contrasts with prior research which has debiased \emph{representations} of text by removing dimensions of prejudice from word embeddings \cite{bolukbasi2016man,gonen2019lipstick,ethayarajh2019understanding} and the hidden states of predictive models \cite{zhao2018gender,das2018mitigating}. To avoid overloading the definition of ``debias,'' we  refer to our kind of text debiasing as \emph{neutralizing} that text. Figure \ref{figure:cnn} gives an example. 

\begin{table*}[ht]
\small
\centering
\begin{tabular}{lll}
\textbf{Source}                                     & \textbf{Target}        & \textbf{Subcategory}                             \\ \hline \hline

A new downtown is being developed which &  A new downtown is being developed which & Epistemological  \\ [-2pt]
will bring back... & \textbf{which its promoters hope} will bring back.. \\  \hline
The authors' \textbf{expos\'{e}} on nutrition studies & The authors' \textbf{statements} on nutrition studies & Epistemological \\ \hline
He started writing books \textbf{revealing} a vast world conspiracy &
He started writing books \textbf{alleging} a vast world conspiracy &Epistemological\\ \hline
Go is \textbf{the deepest} game in the world. & Go is \textbf{one of the deepest} games in the world. &  Framing \\ \hline
Most of the gameplay is \textbf{pilfered from} DDR. & Most of the gameplay is \textbf{based on} DDR.  & Framing \\ \hline
Jewish forces overcome Arab \textbf{militants}.     & Jewish forces overcome Arab \textbf{forces}.   & Framing     \\ \hline
A lead programmer usually spends & Lead programmers often spend & Demographic \\[-2pt]
\textbf{his career} mired in obscurity.  & \textbf{their careers} mired in obscurity.  &       \\ \hline
The lyrics are about \textbf{mankind}'s perceived idea of hell. & The lyrics are about \textbf{humanity}'s perceived idea of hell. & Demographic \\ \hline
Marriage is a \textbf{holy union} of individuals. & Marriage is a \textbf{personal union} of individuals. & Demographic \\ \hline
\end{tabular}
\vspace*{-4pt}
\caption{Samples from our new corpus. 500 sentence pairs are annotated with ``subcategory'' information (Column 3).}
\label{tab:samples}
\end{table*}

We introduce the Wiki Neutrality Corpus (WNC). This is a new parallel corpus of 180,000 biased and neutralized sentence pairs along with contextual sentences and metadata. The corpus was harvested from Wikipedia edits that were designed to ensure texts had a neutral point of view.  WNC is the first parallel corpus of biased language. 

We also define the task of \textit{neutralizing} subjectively biased text.  This task shares many  properties with tasks like detecting framing or epistemological bias \cite{recasens2013linguistic}, or veridicality assessment/factuality prediction \cite{sauri2009factbank,marneffe12,rudinger18,white18}.  Our new task extends these detection/classification problems into a generation task: generating more neutral text with otherwise similar meaning.

Finally, we propose a pair of novel sequence-to-sequence algorithms for this neutralization task. Both methods leverage denoising autoencoders and a token-weighted loss function. An interpretable and controllable \textsc{modular} algorithm breaks the problem into (1) detection and (2) editing, using (1) a BERT-based detector to explicitly identify problematic words, and (2) a novel \emph{join  embedding} through which the detector can modify an editors' hidden states. This paradigm advances an important human-in-the-loop approach to bias understanding and generative language modeling. 
Second, an easy to train and use but more opaque \textsc{concurrent} system uses a BERT encoder to identify subjectivity as part of the generation process.

 Large-scale human evaluation suggests that while not without flaws, our algorithms can identify and reduce bias in encyclopedias, news, books, and political speeches, and do so better than state-of-the-art style transfer and machine translation systems. This work represents an important first step towards automatically managing bias in the real world. We release data and code to the public.\footnote{\url{https://github.com/rpryzant/neutralizing-bias}}

\section{Wiki Neutrality Corpus (WNC)}
\label{section:corpus}

The Wiki Neutrality Corpus  consists of aligned sentences \emph{pre} and \emph{post}-neutralization by English Wikipedia editors (Table \ref{tab:samples}). 
We used regular expressions to crawl 423,823 Wikipedia revisions between 2004 and  2019  where  editors  provided  NPOV-related  justification \cite{zanzotto2010expanding,recasens2013linguistic,yang2017identifying}. To maximize the precision of bias-related changes, we ignored revisions where
\begin{itemize}[noitemsep]
    \item More than a single sentence was changed.
    \item Minimal edits (character Levenshtein distance $<$ 4).
    \item Maximal edits (more than half of the words changed).
    \item Edits where more than half of the words were proper nouns.
    \item Edits that fixed spelling or grammatical errors.
    \item Edits that added references or hyperlinks.
    \item Edits that changed non-literary elements like tables or punctuation.
\end{itemize}

We align sentences in the \emph{pre} and \emph{post} text by computing a sliding window (size $k = 5$) of pairwise BLEU \cite{papineni2002bleu} between sentences and matching sentences with the biggest score \cite{faruqui2018wikiatomicedits,tiedemann2008synchronizing}. Last, we discarded pairs whose length ratios were beyond the 95th percentile \cite{pryzant2017jesc}.

\begin{table}[]
\centering
\small
\begin{tabular}{l|llll}
\textbf{Data}          & \textbf{Sentence} & \textbf{Total} & \textbf{Seq length} & \textbf{\# revised} \vspace{-0.1cm}\\ & \textbf{pairs} & \textbf{words} & \textbf{(mean)} & \textbf{words (mean)} \\ \hline \hline
Biased-full & 181,496         & 10.2M         & 28.21                  & 4.05                     \\
Biased-word & 55,503          & 2.8M          & 26.22                  & 1.00                     \\
Neutral       & 385,639         & 17.4M         & 22.58                  & 0.00                    
\end{tabular}
\caption{Corpus statistics.}
\label{table:corpus-stats}
\end{table}

Corpus statistics are given in Table \ref{table:corpus-stats}. The final data are (1) a parallel corpus of 180k biased sentences and their neutral counterparts, and (2) 385k neutral sentences that were adjacent to a revised sentence at the time of editing but were not changed by the editor. 
Note that following \citet{recasens2013linguistic}, the neutralizing experiments in Section \ref{section:experiments} focus on the  subset of WNC where the editor modified or deleted a single word in the source text (``Biased-word'' in Table \ref{table:corpus-stats}).

Table~\ref{tab:samples} also gives a categorization of these sample pairs using a slight extension of the typology of \citet{recasens2013linguistic}. They defined 
\textbf{framing bias} as using subjective words or phrases linked with a particular point of view (using words like \textit{best} or \textit{deepest} or using \textit{pilfered from} instead of \textit{based on}),  and \textbf{epistemological bias} as linguistic features that subtly (often via presupposition) modify the believability of a proposition. We add to their two a third kind of subjectivity bias that also occurs in our data, which we call \textbf{demographic bias}, text with presuppositions about particular genders, races, or other demographic categories (like presupposing that all programmers are male).  

\begin{table}[h!]
\centering
\begin{tabular}{@{}ll@{}}
\textbf{Subcategory}            & \textbf{Percent} \\ \hline \hline
Epistemological & 25.0   \\
Framing         & 57.7   \\
Demographic     & 11.7   \\
Noise           & 5.6     \\ 
\end{tabular}
\caption{Proportion of bias subcategories in Biased-full.}
\label{tab:bias-types}
\end{table}

The dataset does not include labels for these categories, but we  hand-labeled a random sample of 500  examples to estimate the distribution of the 3 types.
Table \ref{tab:bias-types} shows that while framing bias is most common, all types of bias are represented in the data, including instances of demographic bias.

\subsection{Dataset Properties}
\label{subsection:properties}
We take a closer look at WNC to identify characteristics of subjective bias on Wikipedia.


\textbf{Topic.} We use the Wikimedia Foundation's categorization models \cite{asthana2018few} to bucket articles from WNC into a 44-category ontology,\footnote{\url{https://en.wikipedia.org/wiki/Wikipedia:WikiProject_Council/Directory}} then compare the  proportions of NPOV-driven edits across categories. Subjectively biased edits are  most prevalent in \emph{history}, \emph{politics}, \emph{philosophy}, \emph{sports}, and \emph{language} categories. They are  least prevalent in the \emph{meteorology}, \emph{science}, \emph{landforms}, \emph{broadcasting}, and \emph{arts} categories. This suggests that there is a relationship between a text's topic and the realization of bias. We use this observation to guide our model design in Section \ref{subsection:tagger}.

\textbf{Tenure.}
We group editors into ``newcomers'' (less than a month of experience) and ``experienced'' (more than a month). We find that newcomers are less likely to perform neutralizing edits (15\% in WNC) compared to other edits (34\% in a random sample of 685k edits). This difference is significant ($\tilde{\chi}^2$ p $=$ 0.001), suggesting the complexity of neutralizing text is typically reserved for more senior editors, which helps explain the performance of human evaluators in Section \ref{subsubsection:detection}.

\section{Methods for Neutralizing Text}

We propose the task of neutralizing text, in which the algorithm is given an input sentence and must produce an output sentence whose meaning is as similar as possible to the input but with the subjective bias removed.

We propose two algorithms for this task, each with its own benefits.  A \textsc{modular} algorithm enables human control and interpretability. A \textsc{concurrent} algorithm is simple to train and operate.

We adopt the following notation:
\vspace*{-5pt}\begin{itemize}
    \item $\mathbf{s} = [w^s_1, ..., w^s_n]$ is a \emph{source sequence} of subjectively biased text. \vspace*{-4pt}
    \item $\mathbf{t} = [w^t_1, ..., w^t_m]$ is a \emph{target sequence} and the neutralized version of $\mathbf{s}$.
\end{itemize}

\subsection{MODULAR}
The first algorithm we are proposing is called \textsc{modular}. It has two stages: BERT-based detection and LSTM-based editing. We pretrain a model for each stage and then combine them into a joint system for end-to-end fine tuning on the overall neutralizing task. We proceed to describe each module.  

\subsubsection{Detection Module}  
\begin{figure}[t]
\includegraphics[width=1\linewidth]{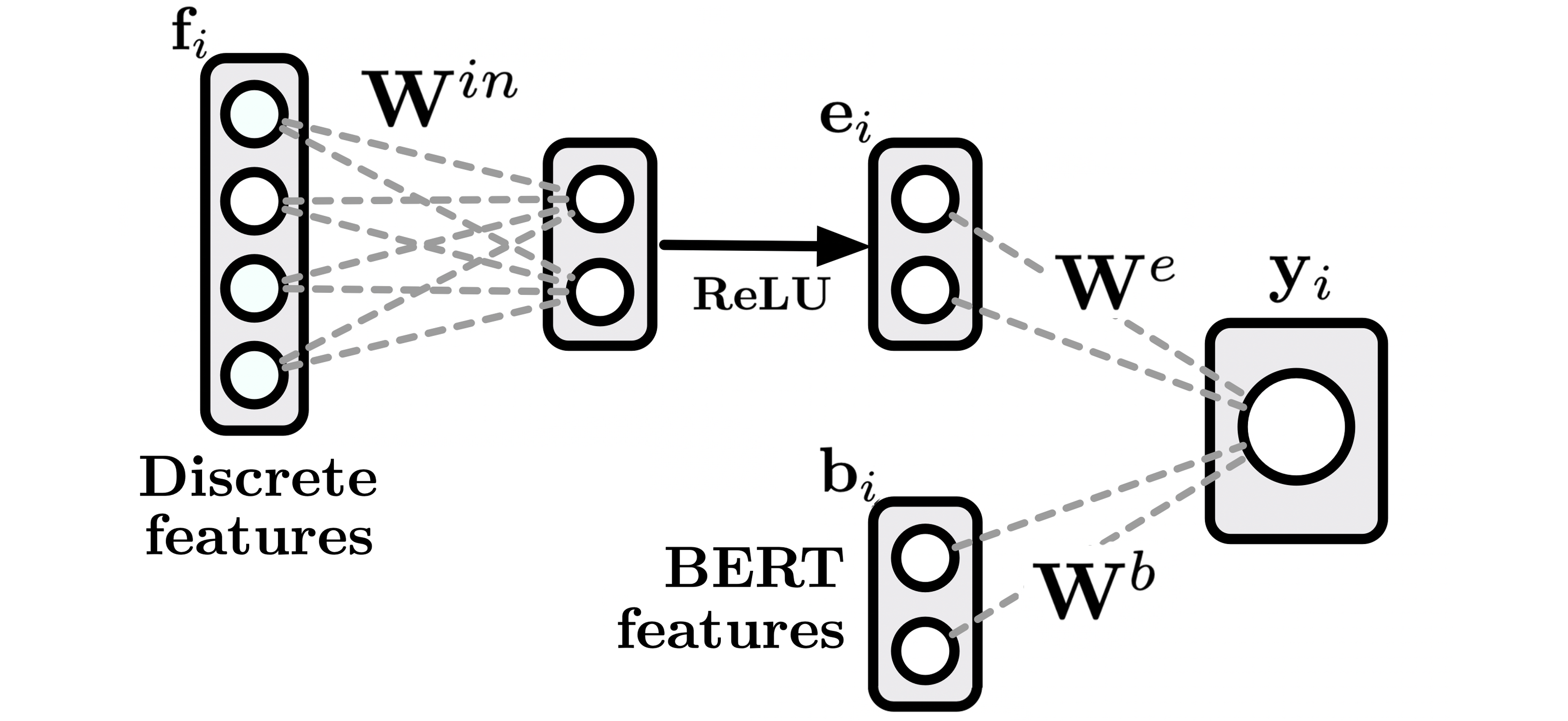}
\caption{The detection module uses discrete features $\mathbf{f}_i$ and BERT embedding $\mathbf{b}_i$ to calculate logit $y_i$.}
\label{figure:tagger}
\end{figure}

The detection module is a neural sequence tagger that estimates $p_i$, the probability that each input word $w^s_i$ is subjectively biased (Figure \ref{figure:tagger}). 
\label{subsection:tagger}

\textbf{Module description.} Each $p_i$ is calculated according to 
\begin{align}
    \label{equation:tagger}
    p_i &= \sigma( \mathbf{b}_i\ \mathbf{W}^{b} + \mathbf{e}_i\ \mathbf{W}^{e} + b)
\end{align}

\begin{itemize}
	\item $\mathbf{b}_i \in \mathcal{R}^{b}$ represents $w^s_i$'s semantic meaning. It is a contextualized word vector produced by BERT, a transformer encoder that has been pre-trained as a masked language model \cite{devlin2018bert}. To leverage the bias-topic relationship uncovered in Section \ref{subsection:properties}, we prepend a token indicating an article's topic category (\texttt{<arts>}, \texttt{<sports>}, etc) to $\mathbf{s}$. The word vectors for these tokens are learned from scratch. 
	
	\item $\mathbf{e}_i$ represents expert features of bias proposed by \cite{recasens2013linguistic}: \vspace*{-4pt}
		\begin{align}
		    \mathbf{e_i} = ReLU(\mathbf{f}_i\ \mathbf{W}^{in})
		\end{align}
		$\mathbf{W}^{in} \in \mathcal{R}^{f \times h}$ is a matrix of learned parameters, and $\mathbf{f}_i$ is a vector of discrete features\footnote{ Such as lexicons of hedges, factives, assertives,  implicatives, and subjective words; see  code release.}. 

	\item $\mathbf{W}^{b} \in \mathcal{R}^{b}$, $\mathbf{W}^{e} \in \mathcal{R}^{h}$, and $b \in \mathcal{R}$ are learnable parameters. 
\end{itemize}
\textbf{Module pre-training.} We train this module using diffs\footnote{\url{https://github.com/paulgb/simplediff}} between the source and target text. A label $p^*_i$ is 1 if $w^s_i$ was deleted or modified as part of the neutralizing process. A label is 0 if the associated word was unchanged during editing, i.e. it occurs in both the source and target text. The loss is calculated as the average negative log likelihood of the labels:
\begin{align*}
   \mathcal{L} = -\ \frac{1}{ n }\sum_{i=1}^n  \Big[ p^*_i \log p_i + (1 - p^*_i) \log (1 - p_i) \Big]
\end{align*}

\subsubsection{Editing Module}
\label{subsection:editor}
The editing module takes a subjective source sentence $\mathbf{s}$ and is trained to edit it into a more neutral compliment $\mathbf{t}$. 

\textbf{Module description.} 
This module is based on a sequence-to-sequence neural machine translation model  \cite{luong2015effective}. A bi-LSTM encoder turns $\mathbf{s}$ into a sequence of hidden states $\mathbf{H} = (\mathbf{h}_1, ..., \mathbf{h}_n)$ \cite{hochreiter1997long}. Next, an LSTM decoder generates text one token at a time by repeatedly attending to $\mathbf{H}$ and producing probability distributions over the vocabulary. We also add two mechanisms from the summarization literature \cite{see2017get}. The first is a copy mechanism,  where the model's final output for timestep $i$ becomes a weighted combination of the predicted vocabulary distribution and attentional distribution from that timestep. The second is a coverage mechanism which incorporates the sum of previous attention distributions into the final loss function to discourage the model from re-attending to a word and repeating itself. 

\textbf{Module pre-training.} 
We pre-train the decoder as a language model of neutral text using the \emph{neutral} portion of WNC (Section \ref{section:corpus}). Doing so expresses a data-driven prior about how target sentences should read.
We accomplish this with a denoising autoencoder objective \cite{hill2016learning} and maximizing the conditional log probability of reconstructing a sequence $\mathbf{x}$ from a \emph{corrupted} version of itself $\widetilde{\mathbf{x}}$ using noise model $C$ ($\log p(\mathbf{x} \vert \widetilde{\mathbf{x}})$ where $\widetilde{\mathbf{x}} = C(\mathbf{x})$).

Our $C$ is similar to \cite{lample2018phrase}. We slightly shuffle $\mathbf{x}$ such that $x_i$'s index in $\widetilde{\mathbf{x}}$ is randomly selected from $[i - k, i + k]$. We then drop words with probability $p$. For our experiments, we set $k = 3$ and $p = 0.25$. 

\begin{figure}[]
\includegraphics[width=1\linewidth]{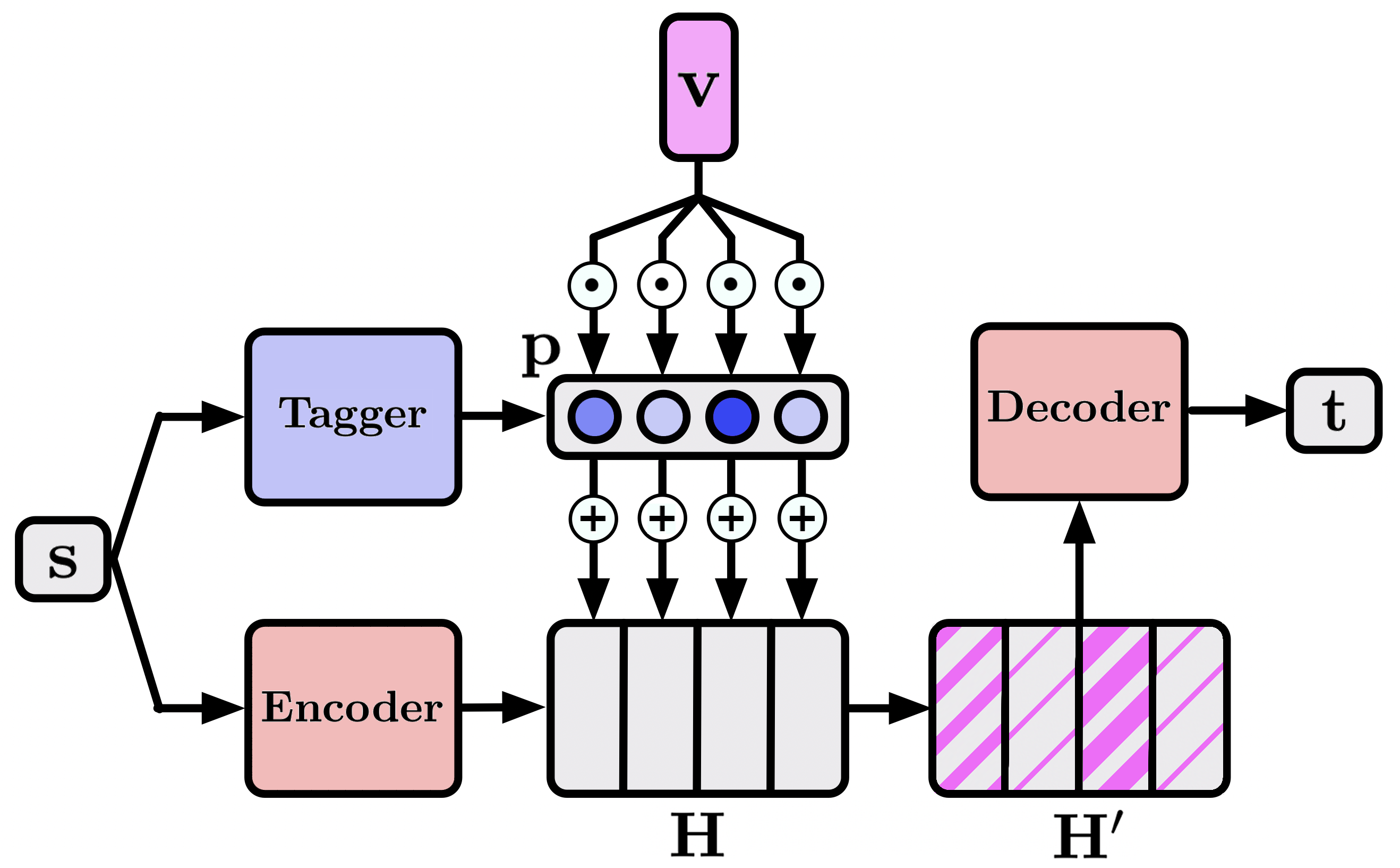}
\caption{The \textsc{modular} system uses \emph{join embedding} $\mathbf{v}$ to reconcile the detector's predictions with an encoder-decoder architecture. The greater a word's probability, the more of $\mathbf{v}$ is mixed into that word's hidden state.}
\label{figure:editor}
\end{figure}

\subsubsection{Final System} 
\label{subsubsection:modular}
Once the detection and editing modules have been pre-trained, we join them and fine-tune together as an end to end system for translating $\mathbf{s}$ into $\mathbf{t}$. 

This is done with a novel \emph{join embedding} mechanism that lets the detector control the editor (Figure \ref{figure:editor}). The join embedding is a vector $\mathbf{v} \in \mathcal{R}^h$ that we add to each encoder hidden state in the editing module. This operation is gated by the detector's output probabilities $\mathbf{p} = (p_1, ..., p_n)$. Note that the same $\mathbf{v}$ is applied across all timesteps.
\vspace*{-4pt}
\begin{align}
    \label{eq:join-embedding}
    \mathbf{h}'_i&= \mathbf{h}_i + p_i \cdot \mathbf{v} 
\end{align}

We proceed to condition the decoder on the new hidden states $\mathbf{H}' = (\mathbf{h'}_1, ..., \mathbf{h}'_n)$ which have varying amounts of $\mathbf{v}$ in them. Intuitively, $\mathbf{v}$ is enriching the hidden states of words that the detector identified as subjective. This tells the decoder what language should be changed and what is safe to be be copied during the neutralization process. 

Error signals are allowed to flow backwards into both the encoder and detector, creating an end-to-end system from the two modules. To fine-tune the parameters of the joint system, we use a token-weighted loss function that scales  the loss on neutralized words (i.e. words unique to $\mathbf{t}$) by a factor of  $\alpha$:
\vspace*{-4pt}
\begin{align*}
    \mathcal{L}(s, t) &= - \sum_{i=1}^m \lambda(w^t_i, s) \log p(w^t_i \vert s, w^t_{<i}) + c \\
    \lambda(w^t_i, s) &= \left\{
                \begin{array}{ll}
                  \alpha\ :\ w^t_i \not \in s\\
                  1\ :\ \mathrm{otherwise}
                \end{array}
        \right.
\end{align*}
Note that $c$ is a term from the coverage mechanism (Section \ref{subsection:editor}). We use $\alpha = 1.3$ in our experiments. Intuitively, this loss function incorporates an inductive bias of the neutralizing  process: the source and target have a high degree of lexical similarity but the goal is to learn the structure of their \emph{differences}, not simply copying words into the output (something a pre-trained autoencoder should already have knowledge of). This loss function is related to previous work on grammar correction \cite{junczys2018approaching}, and cost-sensitive learning \cite{zhou2006training}.

\subsection{CONCURRENT}

Our second algorithm takes the problematic source $\textbf{s}$ and directly generates a neutralized $\mathbf{\hat{t}}$. While this renders the system easier to train and operate, it limits interpretability and controllability. 

\textbf{Model description}. The \textsc{concurrent} system is an encoder-decoder neural network. The encoder is BERT. The decoder is the same as that of Section \ref{subsection:editor}: an attentional LSTM with copy and coverage mechanisms. The decoder's inputs are set to: 
\begin{itemize}
	\item Hidden states $\mathbf{H} = \mathbf{W}^H\ \mathbf{B}$, where $\mathbf{B} = (\mathbf{b}_1, ..., \mathbf{b}_{n}) \in \mathcal{R}^{b \times n}$ is the BERT-embedded source and $\mathbf{W}^H \in \mathcal{R}^{h \times b}$ is a matrix of learned parameters.
	\item Initial states $\mathbf{c}_0 =  \mathbf{W}^{c0}\ \sum \mathbf{b}_i / n$ and $\mathbf{h_0} = \mathbf{W}^{h0}\ \sum \mathbf{b}_i / n$. $\mathbf{W}^{c0} \in \mathcal{R}^{h \times b}$ and $\mathbf{W}^{h0} \in \mathcal{R}^{h \times b}$ are learned matrices.
\end{itemize}

\textbf{Model training}.
The \textsc{concurrent} model is pre-trained with the same autoencoding procedure described in Section \ref{subsection:editor}. It is then fine-tuned as a subjective-to-neutral translation system with the same loss function described in Section \ref{subsubsection:modular}.

\begin{table*}[]
\centering
\begin{tabular}{l|ll|lll}
\textbf{Method}                                                  & \textbf{BLEU} & \textbf{Accuracy} & \textbf{Fluency} & \textbf{Bias} & \textbf{Meaning} \\ \hline \hline
Source Copy                                             & 91.33 & 0.00  &   -     &        -      &    -       \\ \hline
Detector (always delete biased word)                                  & 92.43* & 38.19* &  -0.253* &   -0.324*         &      1.108*            \\ \hline
Detector (predict substitution from biased word)                                 & 92.51 & 36.57* &  -0.233*     &   -0.327*        &      1.139*         \\ \hline
Delete Retrieve (ST) \cite{li2018delete}                & 88.46* & 14.50* &    -0.209*      &   -0.456*     &     1.294*           \\ \hline
Back Translation (ST) \cite{prabhumoye2018style}        & 84.95* & 9.92* &   -0.359*     &       -0.390*  &   1.126*           \\ \hline
Transformer (MT) \cite{vaswani2017attention}           & 86.40* & 24.34* &  -0.259*     &    -0.458*      &    0.905*              \\ \hline
Seq2Seq (MT) \cite{luong2015effective}                 & 89.03* & 23.93 &  -0.423*        &   -0.436*       &    1.294*          \\  \hline  
Base                                                    & 89.13 & 24.01 &   -      &    -          &    -       \\  
\emph{+ loss}                                             & 90.32* & 24.10 &  -       &  -            &   -        \\  
\emph{+ loss + pretrain }                                      & 92.89* & 34.76* &   -      &         -     &     -      \\
\emph{+ loss + pretrain + detector} (\textsc{modular})      & 93.52* & \textbf{45.80}* &   -0.078     &   \textbf{-0.467}*    &     0.996*             \\ 
\emph{+ loss + pretrain + BERT} (\textsc{concurrent})   & \textbf{93.94} & 44.87 &   \textbf{0.132}     &    -0.423*     &  \textbf{0.758}*              \\ \hline  
Target copy                                            & 100.0 & 100.0     & -0.077       &  -0.551*     &    1.128*              \\ 
\end{tabular}
\caption{Bias neutralization performance. ST indicates a style transfer system. MT indicates a machine translation system.  For quantitative metrics, rows with asterisks are significantly different than the preceding row. For qualitative metrics, rows with asterisks are significantly different from zero. Higher is preferable for \emph{fluency}, while lower is preferable for \emph{bias} and \emph{meaning}.}
\label{tab:editing}
\end{table*}

\section{Experiments}
\label{section:experiments}

\subsection{Experimental Protocol}
\label{sub:setup}

\textbf{Implementation.} We implemented nonlinear models with Pytorch \cite{paszke2017automatic} and optimized using Adam \cite{kingma2014adam} as configured in \cite{devlin2018bert} with a learning rate of 5e-5. We used a batch size of 16. All vectors were of length $h = 512$ unless otherwise specified. We use gradient clipping with a maximum gradient norm of 3 and a dropout probability of 0.2 on the inputs of each LSTM cell \cite{srivastava2014dropout}. We initialize the BERT component of the tagging module with the publicly-released \texttt{bert-base-uncased} parameters. All other parameters were uniformly initialized in the range $[-0.1,\ 0.1]$.

\textbf{Procedure.} Following \citet{recasens2013linguistic}, we train and evaluate our system on a subset of WNC where the editor changed or deleted a single word in the source text. This yielded 53,803 training pairs (about a quarter of the WNC), from which we sampled 700 development and 1,000 test pairs. For fair comparison, we gave our baselines additional access to the 385,639 \emph{neutral} examples when possible. We pretrained the tagging module for 4 epochs. We pretrained the editing module on the \emph{neutral} portion of our WNC for 4 epochs. The joint system was trained on the same data as the tagger for 25,000 steps (about 7 epochs).  We perform interference using beam search and a beam width of 4. All computations were performed on a single NVIDIA TITAN X GPU; training the full system took approximately 10 hours.

\textbf{Evaluation.} We evaluate our models according to five metrics. BLEU \cite{papineni2002bleu} and accuracy (the proportion of decodings that exactly matched the editors changes) are quantitative. We report statistical significance with bootstrap resampling and a 95\% confidence level  \cite{koehn2004statistical,efron1994introduction}.

We also hired fluent English-speaking crowdworkers on Amazon Mechanical Turk for qualitative evaluation. Workers were shown the \citet{recasens2013linguistic} and Wikipedia definition of a ``biased statement'' and six example sentences, then subjected to a five-question qualification test where they had to identify subjectivity bias. Approximately half of the 30,000 workers who took the qualification test passed. Those who passed were asked to compare pairs of original and edited sentences (not knowing which was the original) along three criteria: fluency, meaning preservation, and bias. Fluency and bias were evaluated on a Semantic Differential scale from -2 to 2. Here, a semantic differential scale can better evaluate attitude oriented questions with two polarized options (e.g., ``is text A or B more fluent?''). 
Meaning was evaluated on a Likert scale from 0 to 4, ranging from ``identical'' to ``totally different''. Inter-rater agreement was fair to substantial (Krippendorff's alpha of 0.65 for fluency, 0.33 for meaning, and 0.51 for bias)\footnote{\ Rule of thumb: k $<$ 0 ``poor'' agreement, 0 to .2 ``slight'', .21 to .40 ``fair'', .41 to .60 ``moderate'', .61 - .80 ``substantial'', and .81 to 1 ``near perfect'' \cite{gwet2011krippendorff}.}. We report statistical significance with a t-test and 95\% confidence interval.


\subsection{Wikipedia (WNC)}

Results on WNC are presented in Table \ref{tab:editing}. In addition to methods from the literature we include (1) a BERT-based system which simply predicts and deletes subjective words, and (2) a system which predicts replacements (including deletion) for subjective words directly from their BERT embeddings. All methods appear to successfully reduce bias according to the human evaluators. However, many methods appear to lack fluency. 
Adding a token-weighted loss function and pretraining the decoder help the model's coherence according to BLEU and accuracy. Adding the detector (\textsc{modular}) or a BERT encoder (\textsc{concurrent}) provide additional benefits. The proposed models retain the strong effects of systems from the literature while also producing target-level fluency on average. Our results suggest there is no clear winner between our two proposed systems. \textsc{modular} is better at reducing bias and has higher accuracy, while \textsc{concurrent} produces more fluent responses, preserves meaning better, and has higher BLEU. 

\begin{table}[]
\centering
\begin{tabular}{l|lll}
\textbf{Metric}  & \textbf{Fluency} & \textbf{Bias} & \textbf{Meaning}  \\ \hline \hline
BLEU     & 0.65   & 0.34   & 0.16 \\ 
Accuracy & 0.56   & 0.52   & 0.20\\ 
\end{tabular}
\caption{Spearman correlation ($R^2$) between quantitative and qualitative metrics.}
\label{tab:correlations}
\end{table}

Table \ref{tab:correlations} indicates that BLEU is more correlated with fluency but accuracy is more correlated with subjective bias reduction. The weak association between BLEU and human evaluation scores is corroborated by other research \cite{chaganty2018price,mir2019evaluating}. We conclude that neither automatic metric is a true substitute for human judgment. 

\subsection{Real-world Media}

To demonstrate the efficacy of the proposed methods on subjective bias in the wild, we perform inference on three out-of-domain datasets (Table \ref{tab:other-results}). We prepared each dataset according to the same procedure as WNC (Section \ref{section:corpus}). After inference, we enlisted 1800 raters to assess the quality of 200 randomly sampled datapoints.
Note that for partisan datasets we sample an equal number of examples from ``conservative'' and ``liberal'' sources. These data are:

\vspace*{-4pt}
\begin{itemize}
\itemsep 1pt
\item The Ideological Books Corpus (IBC) consisting of partisan books and magazine articles \cite{sim2013measuring,iyyer2014political}. 
\vspace*{-4pt}
\item Headlines of partisan news articles identified as biased according to \url{mediabiasfactcheck.com}.
\vspace*{-4pt}
\item Sentences from the campaign speeches of a prominent politician (United States President Donald Trump).\footnote{Transcripts from \url{www.kaggle.com/binksbiz/mrtrump}}  We filtered out dialog-specific artifacts (interjections, phatics, etc) by removing all sentences with less than 4 tokens before sampling a test set. 
\end{itemize}

Overall, while \textsc{modular} does a better job at reducing bias, \textsc{concurrent} appears to better preserve the meaning and fluency of the original text. We conclude that the proposed methods, while imperfect, are capable of providing useful suggestions for how subjective bias in real-world news or political text can be reduced.

\begin{table}[]
\begin{tabular}{llll}
\multicolumn{4}{c}{\textbf{IBC Corpus}}                          \\ \hline \hline
\multicolumn{1}{l|}{Method}           & Fluency & Bias  & Meaning\\ \hline
\multicolumn{1}{l|}{\textsc{modular}}  &  -0.041      &  -0.509*    &  0.882*       \\
\multicolumn{1}{l|}{\textsc{concurrent}} &  -0.001      &  -0.184      &  0.501*      \\ \hline \hline
\multicolumn{1}{l|}{Original} & \multicolumn{3}{l}{\small \textbf{Activists} have filed a lawsuit...} \\
\multicolumn{1}{l|}{\textsc{modular}} & \multicolumn{3}{l}{\small \textbf{Critics of it} have filed a lawsuit...} \\
\multicolumn{1}{l|}{\textsc{concurrent}} & \multicolumn{3}{l}{\small \textbf{Critics} have filed a lawsuit...}
\\  \\
\multicolumn{4}{c}{\textbf{News Headlines}}                         \\ \hline \hline
\multicolumn{1}{l|}{Method}           & Fluency & Bias & Meaning \\ \hline
\multicolumn{1}{l|}{\textsc{modular}}  &   -0.46*      &  -0.511*     &   1.169*     \\
\multicolumn{1}{l|}{\textsc{concurrent}} &   -0.141*     &  -0.393*     &  0.752*       \\ \hline \hline
\multicolumn{1}{l|}{Original} & \multicolumn{3}{l}{\small Zuckerberg \textbf{claims} Facebook can...} \\
\multicolumn{1}{l|}{\textsc{modular}} & \multicolumn{3}{l}{\small Zuckerberg \textbf{stated} Facebook can...} \\
\multicolumn{1}{l|}{\textsc{concurrent}} & \multicolumn{3}{l}{\small Zuckerberg \textbf{says} Facebook can...} \\ \\
\multicolumn{4}{c}{\textbf{Trump Speeches}}                         \\ \hline \hline
\multicolumn{1}{l|}{Method}           & Fluency & Bias  & Meaning\\ \hline
\multicolumn{1}{l|}{\textsc{modular}}  &  -0.353*      &  -0.563*     &  1.052*        \\
\multicolumn{1}{l|}{\textsc{concurrent}} &  -0.117      &  -0.127    &  0.757*     \\\hline \hline
\multicolumn{1}{l|}{Original} & \multicolumn{3}{l}{\small This includes \textbf{amazing} Americans like...} \\
\multicolumn{1}{l|}{\textsc{modular}} & \multicolumn{3}{l}{\small This includes Americans like...} \\
\multicolumn{1}{l|}{\textsc{concurrent}} & \multicolumn{3}{l}{\small This includes \textbf{some} Americans like...} 
\end{tabular}
\caption{Performance on out-of-domain datasets. Higher is preferable for \emph{fluency}, while lower is preferable for \emph{bias} and \emph{meaning}. Rows with asterisks are significantly different from zero}
\label{tab:other-results}
\end{table}

\section{Error Analysis}
To better understand the limits of our models and the proposed task of bias neutralization, we randomly sample 50 errors produced by our models on the Wikipedia test set and bin them into the following categories:


\begin{itemize}
    \item \textbf{No change.} The model failed to remove or change the source sentence.
    \item \textbf{Bad change.} The model modified the source but introduced an edit which failed to match the ground-truth target (i.e. the Wikipedia editor's change).
    \item \textbf{Disfluency.} Errors in language modeling and text generation.
    \item \textbf{Noise.} The datapoint is noisy and the target text is not a neutralized version of the source.
\end{itemize}

\begin{table}[h!]
\centering
\begin{tabular}{l|ll}
\textbf{Error Type} & \textbf{Proportion (\%)} & \textbf{Valid (\%)} \\ \hline \hline
No change  & 38                & 0            \\
Bad change & 42                & 80           \\
Disfluency & 12                & 0            \\
Noise      & 8                 & 87         
\end{tabular}
\caption{Distribution of model errors on the Wikipedia test set. We also give the percent of errors that were valid neutralizations of the source despite failing to match the target sentence.}
\label{tab:error-prop}
\end{table}

The distribution of errors is given in Table \ref{tab:error-prop}. Most errors are due to the subtlety and complexity of language understanding required for bias neutralization, rather than the generation of fluent text. These challenges are particularly pronounced for neutralizing edits that involve the replacement of factive and assertive verbs. As column 2 shows, a large proportion of the errors, though disagreeing with the edit written by the Wikipedia editors, nonetheless succeeded in neutralizing the source.

Examples of each error type are given in Table \ref{tab:error-examples} (two pages away). As the examples show, our models have have a tendency to simply remove words instead of finding a good replacement.

\begin{table}[htb]
\centering
\begin{tabular}{l|l}
\textbf{Method}                       & \textbf{Accuracy}  \\ \hline \hline
Linguistic features              &    0.395*        \\ \hline
Bag-of-words                        &   0.584*          \\
\it{+Linguistic features}      &   0.617          \\ 
\ \ \ \ \ (Recasens, 2013)      &              \\ \hline
BERT                         &     0.744*     \\
\it{ +Linguistic features} &     0.752     \\ 
\it{ +Linguistic features + Category} &    \textbf{0.759}         \\ 
\ \ \ \ \ \ \ (\textsc{modular} detector) &             \\ \hline
\textsc{concurrent} encoder &    0.745    \\ \hline
Human                        &     0.543*         \\ 
\end{tabular}
\caption{Performance of various bias detectors. Rows with asterisks are statistically different than the preceding row.}
\label{tab:tagging}
\end{table}

\begin{table*}[htb]
\centering
\small
\begin{tabular}{ll}
\multicolumn{1}{l|}{\textbf{Error Type}} & \textbf{Source, Output, then Target} \\ \hline \hline
\multicolumn{1}{l|}{No change}  & Existing hot-mail accounts were \textbf{upgraded} to outlook.com on April 3, 2013. \\ 
\multicolumn{1}{l|}{}           & Existing hot-mail accounts were \textbf{upgraded} to outlook.com on April 3, 2013. \\
\multicolumn{1}{l|}{}           & Existing hot-mail accounts were \textbf{changed} to outlook.com on April 3, 2013. \\ \hline \hline
\multicolumn{1}{l|}{Bad change} & His \textbf{exploitation} of leased labor began in 1874 and continued until his death in 1894...                 \\
\multicolumn{1}{l|}{}           & His \textbf{actions} of leased labor began in 1874 and continued until his death in 1894...   \\
\multicolumn{1}{l|}{}           & His \textbf{use} of leased labor began in 1874 and continued until his death in 1894...   \\ \hline \hline
\multicolumn{1}{l|}{Disfluency} & Right before stabbing a cop, flint attacker shouted one thing that \textbf{proves} terrorism is still here. \\ 
\multicolumn{1}{l|}{}          & Right before stabbing a cop, flint attacker shouted one thing that \textbf{may may} terrorism is still here.  \\
\multicolumn{1}{l|}{}          & Right before stabbing a cop, flint attacker shouted one thing that \textbf{may prove} terrorism is still here.  \\ \hline \hline
\multicolumn{1}{l|}{Noise} & ...then whent to war with him in the Battle of \textbf{Bassorah}, and ultimately left that battle. \\ 
\multicolumn{1}{l|}{}          &  ...then whent to war with him in the Battle of \textbf{Bassorah}, and ultimately left that battle.  \\
\multicolumn{1}{l|}{}  & ...then whent to war with him in the Battle of \textbf{the Camel}, and ultimately left that battle.  \\ \\  \\ 

\multicolumn{1}{l|}{\textbf{Revised Word}} & \textbf{Source, Output, then Target} \\ \hline \hline
\multicolumn{1}{l|}{Magnificent}  & After a dominant performance, Joshua...with a \textbf{magnificent} seventh-round knockout win. \\ 
\multicolumn{1}{l|}{}           & After a dominant performance, Joshua...with a seventh-round knockout win. \\
\multicolumn{1}{l|}{}           & After  a  dominant  performance, Joshua...with a seventh-round knockout win. \\ \hline \hline
\multicolumn{1}{l|}{Dominant} & Jewish history is...interacted with other \textbf{dominant} peoples, religions and cultures.                 \\
\multicolumn{1}{l|}{}           & Jewish history is...other peoples, religions and cultures.   \\
\multicolumn{1}{l|}{}           & Jewish history is...other peoples, religions and cultures.  \\ \\ \\ 

 \multicolumn{1}{l|}{\textbf{Selected Word}} & \textbf{Output}  \\ \hline \hline
\multicolumn{1}{l|}{ \emph{(input)}} & In recent years, the term has often been \textbf{misapplied} to those who are \textbf{merely} clean-cut.        \\
\multicolumn{1}{l|}{merely} & In recent years, the term has often been \textbf{misapplied} to those who are clean-cut.     \\
 \multicolumn{1}{l|}{misapplied}  & In recent years, the term has often been \emph{shown} to those who are \textbf{merely} clean-cut.  \\ \hline \hline
 \multicolumn{1}{l|}{\emph{(input)}} & He was responsible for the \textbf{assassination} of Carlos Marighella, and for the Lapa \textbf{massacre}. \\
 \multicolumn{1}{l|}{assassination} & He was responsible for the \emph{killing} of Carlos Marighella, and for the Lapa \textbf{massacre}. \\
 \multicolumn{1}{l|}{massacre} & He was responsible for the assassination of Carlos Marighella, and for the Lapa \emph{incident}. \\ \hline \hline

\multicolumn{1}{l|}{ \emph{(input)}} & Paul Ryan \textbf{desperately} searches for a new focus amid Russia \textbf{scandal}. \\
 \multicolumn{1}{l|}{desperately} & Paul Ryan searches for a new focus amid Russia \textbf{scandal}. \\
 \multicolumn{1}{l|}{scandal} & Paul Ryan \textbf{desperately} searches for a new focus amid Russia. \\  \\  
\end{tabular}
\vspace{-0.2cm}
\caption{\textbf{Top}: examples of model errors from each error category. \textbf{Middle}: the model treats words differently based on their context; in this case, ``dominant'' is ignored when it accurately describes an individual's winning performance, but deleted when it describes a group of people in arbitrary comparison. \textbf{Bottom}: the \textsc{modular} model can sometimes be controlled, for example by selecting words to change, to correct errors or otherwise change the model's behavior.}
\label{tab:error-examples}
\end{table*}

\section{Algorithmic Analysis}

We proceed to analyze our algorithm's ability to detect and categorize bias as well as the efficacy of the proposed join embedding.

\subsection{Detecting Subjectivity}
\label{subsubsection:detection}

Identifying subjectivity in a sentence (explicitly or implicitly) is prerequisite to neutralizing it. We accordingly evaluate our model's (and 3,000 crowdworker's) ability to detect subjectivity using the procedure of \citet{recasens2013linguistic}. We use the same 50k training examples as Section \ref{section:experiments} (Table \ref{tab:tagging}). For each sentence, we select the word with the highest predicted probability and test whether that word was in fact changed by the editor. The proportion of correctly selected words is the system's ``accuracy''. Results are given in Table \ref{tab:tagging}.

Note that \textsc{concurrent} lacks an interpretive window into its detection behavior, so we estimate an upper bound on the model's detection abilities by (1) feeding the encoder's hidden states into a fully connected + softmax layer that predicts the probability of a token being subjectively biased, and (2) training this layer as a sequence tagger according to the procedure of Section \ref{subsection:tagger}. 

The low human performance can be attributed to the difficulty of identifying bias. Issues of bias are typically reserved for senior Wikipedia editors (Section \ref{subsection:properties}) and untrained workers performed worse (37.39\%) on the same task in \cite{recasens2013linguistic} (and can struggle on other tasks requiring linguistic knowledge \cite{callison2009fast}). \textsc{concurrent}'s encoder, which is architecturally identical to BERT, had similar performance to a stand-alone BERT system. The linguistic and category-related features in the \textsc{modular} detector gave it slight leverage over the plain BERT-based models.

\subsection{Join Embedding}

We continue by analyzing the abilities of the proposed join embedding mechanism.

\subsubsection{Join Embedding Ablation}
The join embedding combines two separately pretrained models through a gated embedding instead of the more traditional practice of stripping off any final classification layers and concatenating the exposed hidden states \cite{bengio2007greedy}. We ablated the join embedding mechanism by training a new model where the pre-trained detector is frozen and its pre-output hidden states $\mathbf{b}_i$ are concatenated to the encoder's hidden states before decoding. Doing so reduced performance to 90.78 BLEU and 37.57 Accuracy (from the 93.52/46.8 with the join embedding). This suggests learned embeddings can be a high-performance and end-to-end conduit between sub-modules of machine learning systems. 


\subsubsection{Join Embedding Control}

We proceed to demonstrate how the join embedding creates controllability in the neutralization process. Recall that \textsc{modular} relies on a probability distribution $\mathbf{p}$ to determine which words require editing (Equation \ref{eq:join-embedding}). Typically, this distribution comes from the detection module (Section \ref{subsection:tagger}), but we can also feed in user-specified distributions that force the model to target particular words. This can let human advisors correct errors or push the model's behavior towards some desired outcome. We find that the model is indeed capable of being controlled, letting users target specific words for rewording in case they disagree with the model's output or seek recommendations on specific language. However, doing so can also introduce errors into downstream language generation (Table \ref{tab:error-examples}, next page). 

\section{Related Work}

\textbf{Subjectivity Bias.} 
The study of subjectivity in NLP was pioneered by the late Janyce Wiebe and colleagues \cite{bruce1999recognizing,hatzivassiloglou2000effects}.
Several studies develop methods for highlighting subjective or persuasive frames in a text \cite{rashkin2017truth,tsur2015frame}, 
or detecting biased sentences
\cite{hube2018detecting,morstatter2018identifying,yang2017identifying,hube2019neural}
of which the most similar to ours is \citet{recasens2013linguistic}, whose early, smaller version of WNC and logistic regression-based bias detector inspired our study.


\textbf{Debiasing.}
Many scholars have worked on removing demographic prejudice from \emph{meaning representations} \cite[inter alia]{manzini2019black,zhao2017men,zhao2018gender,bordia2019identifying,wang2018adversarial}. 
Such studies begin with identifying a direction or subspace that capture the bias and then removing this bias component to make representations fair across attributes like gender and age \cite{bolukbasi2016man,manzini2019black}.
For instance, \citet{bordia2019identifying} introduced a regularization term for the language model to penalize the projection of the word embeddings onto that gender subspace, while \citet{wang2018adversarial} used adversarial training to squeeze directions of bias out of hidden states.

\textbf{Neural Language Generation.} Several studies propose stepwise or modular procedures for text generation, including sampling from a corpus \cite{guu2018generating} and identifying language ripe for modification \cite{leeftink2019towards}. Most similar to us is \citet{li2018delete} who localize a text's style to a fraction of its words. Our \textsc{modular} detection module performs a similar localization in a soft manner, but our steps are joined by a smooth conduit (the join embedding) instead of discrete logic. There is also work related to our \textsc{concurrent} model. The closest is \citet{dunextending}, where a decoder was attached to BERT for question answering, or \citet{lample2018phrase}, where machine translation systems are initialized to LSTM and Transformer-based language models of the source text. 


\section{Conclusion and Future Work}

The growing presence of bias has marred the credibility of our news, educational systems, and social media platforms. Automatically reducing bias is thus an important new challenge for the Natural Language Processing and Artificial Intelligence community. This work represents a first step in the space. Our results suggest that the proposed models are capable of providing useful suggestions for how to reduce subjective bias in real-world expository writing like news, books, and encyclopedias. Nonetheless our scope was limited to single-word edits, which only constitute a quarter of the edits in our data, and are probably among the simplest instances of bias.
We therefore encourage future work to tackle broader instances of multi-word, multi-lingual, and cross-sentence bias. Another important direction is integrating aspects of fact-checking \cite{mihaylova2018fact}, since a more sophisticated system would be able to know when a presupposition is in fact true and hence not subjective. Finally, our new join embedding mechanism can be applied to other modular neural network architectures.

\section{Acknowledgements}

We thank the Japan-United States Educational Commission (Fulbright Japan) for their generous support.
We thank Chris Potts, Hirokazu Kiyomaru, Abigail See, Kevin Clark, the Stanford NLP Group, and our anonymous reviewers for their thoughtful comments and suggestions.
We gratefully acknowledge support of the DARPA Communicating with Computers (CwC) program
under ARO prime contract no. W911NF15-1-0462 and the NSF via grant IIS-1514268.
Diyi Yang is thankful for support by a grant from Google.

\bibliography{aaai}
\bibliographystyle{aaai}
\end{document}